# Solving Oscillator Ordinary Differential Equations in the Time Domain with High Performance via Soft-constrained Physics-informed Neural Network with Small Data

**Kai-liang Lu[1,2], Yu-meng Su[1]**

[1]College of Information Technology, Shanghai Jian Qiao University, Shanghai, 201306, China
[2]School of Urban Railway Transportation, Shanghai University of Engineering Science, Shanghai 201620, China
[*]Corresponding Author: Kai-liang Lu. Email: lukailiang@163.com

**ABSTRACT:** In many scientific and engineering (e.g., physical, biochemical, medical) practices, data generated through expensive experiments or large-scale simulations, are often sparse and noisy. Physics-informed neural network (PINN) incorporates physical information and knowledge into network topology or computational processes as model priors, with the unique advantage of achieving strong generalization with small data. This study aims to investigate the performance characteristics of the soft-constrained PINN method to solving typical linear and nonlinear ordinary differential equations (ODEs) such as primer, Van der Pol and Duffing oscillators, especially the effectiveness, efficiency, and robustness to noise with minimal data[1]. It is verified that the soft-constrained PINN significantly reduces the need for labeled data. With the aid of appropriate collocation points no need to be labeled, it can predict and also extrapolate with minimal data. First-order and second-order ODEs, no matter linear or nonlinear oscillators, require only one and two training data (containing initial values) respectively, just like classical analytic or Runge-Kutta methods, and with equivalent precision and comparable efficiency (fast training in seconds for scalar ODEs). Moreover, PINN is naturally robust to noisy data, thus with enhanced generalization capabilities. Furthermore, it can conveniently impose a physical law (e.g., conservation of energy) constraint by adding a regularization term to the total loss function, improving the performance to deal with various complexities such as nonlinearity like Duffing. The DeepXDE-based PINN implementation is light code and can be efficiently trained on both GPU and CPU platforms. The mathematical and computational framework of this alternative and feasible PINN method to ODEs, can be easily extended to PDEs, etc.

## 1 Introduction

Differential equations (DEs), such as ODEs and partial DEs (PDEs), are important tools for expressing the laws of nature in science and engineering. The behavior of complex real-world systems can be modeled using these DEs based on different domain-specific assumptions and simplifications [12,22]. The common representative numerical discretization (ND) methods for solving ODEs/PDEs are the Runge-Kutta method [15] and the finite element method (FEM) [1,14], respectively. Great progress has been made in solving DEs by ND methods in order to simulate various types of field problems. Its theoretical foundations are complete and it is interpretable - there are readily available error estimates as well as convergence and stability guarantees, and with the advantage of high efficiency and accuracy.

However, there are still some bottlenecks that severely limit its application: (i) the curse of dimensionality problem, (ii) the mesh generation is still complicated, (iii) there are difficulties in merging experimental data, still not easy to seamlessly incorporate noisy data into traditional ND methods, and (iv) solving inverse problems (e.g., inferring material properties in functional materials or discovering missing physics in reaction transport [9]) is often very expensive. For example, an obvious disadvantage of FEM is that it relies on spatial discretization (spatial meshes plus large polynomial bases) and suffers from the curse of dimensionality. It is already difficult to use in three dimensions, let alone for higher dimensional problems. Moreover, certain nonlinear and non-smooth PDEs are very difficult to discretize. Due to general non-smooth behavior or singularities, they usually need to be solved on a very fine mesh [12].

[1] Code available at: https://github.com/mikelu-shanghai/PINNtoODEwithSmallData

With the explosive growth of available data and computational resources, deep learning have achieved superior human performance in several tasks such as image recognition and chess playing [18,30]. Its success relies foremost on big data but it suffers from overfitting, poor generalization performance and has interpretability issues [6,7]. On the other hand, in real complex scientific and engineering (e.g., physical, biochemical, medical) practices [8,23,32], data are often generated through expensive experiments or large-scale simulations, which are usually sparse and noisy [24]. Sometimes the cost of obtaining data is prohibitively high, and large amounts of data are not even available at all [16,38].

In these small data scenarios, conventional neural networks (NN) lack robustness and do not provide convergence guarantees [27]. We are inevitably facing the challenge of predicting or making decisions with partial information. The good news is that there is a great deal of prior knowledge (i.e., the culmination of previous wisdom) [10,11] in these system modeling cases that has not been fully exploited in conventional NN. Deep NN is highly expressive, thus neglecting to utilize a prior information or knowledge, yet also leading to over-reliance on data and computational power.

The recently developed PINN [27] emphasized on the use of physical information or prior knowledge in the topological structure and learning process of NN. This is a new class of general-purpose function approximators that inherits the strong expressive power of NN. It is able to encode any of the fundamental laws of physics often described by DEs that govern a given data set [22]. PINN has the unique advantage of achieving strong generalization with small data. By enforcing or embedding physics, the neural network model is effectively constrained to a lower dimensional manifold and therefore requires only a small amount of data for training [16]. And these prior knowledge or constraints can yield more interpretable learning methods. PINN is not only capable of interpolation but also extrapolation, and it remains robust to data imperfections [16].

Deep NN methods such as PINN [27] (also deep Ritz method [34], deep Galerkin method [31]) have achieved great success in solving high-dimensional problems and become an alternative solution method for various DEs. They have the potential to overcome some of the challenges faced by the ND methods described above: (i) By utilizing network architecture and automatic differentiation [5], the need for discretization is eliminated, being mesh-free. (ii) Neural networks are able to represent more general functions than finite element bases, can break the curse of dimensionality to some extend [25]. (iii) Although training neural networks (non-convex optimization) may become computationally intensive compared to traditional ND solvers, it is very effective and efficient in evaluating new data samples. Table 1 summarizes the properties of the three types of methods for solving DEs.

**Table 1:** Comparison of the features of the methods for solving differential equations

| Method | Traditional Numerical Discretization (e.g. Runge-Kutta, FEM) | Conventional Neural Network | Physics-informed Neural Network |
|---|---|---|---|
| Feature | ·Complete theoretical foundation, interpretable, readily available error estimates, with convergence and stability guarantees<br>·High efficiency | ·Mesh-free<br>·Breaking the curse of dimensionality to some extend, can cope with high dimensional and nonlinear problems | ·Advantages of conventional NN +<br>·Small data, fewer network parameters, less prone to overfitting, strong generalization capability<br>·Robust to imperfect data and |

| | | |
|---|---|---|
| ·High precision | ·Generalizability to new data | incomplete models |
| | samples | ·Effective and efficient in dealing |
| ·Suffering from curse of dimensionality | | with ill-posed and inverse problems |
| ·Mesh generation is still complex | ·Need big data and strong | ·More interpretable than NN |
| ·Difficult to seamlessly integrate noisy data | computing power | ·Fast training on CPU/GPU |
| ·Excessive cost of solving inverse problem | ·Overfitting | |
| | ·Black box issue | ·Black box issue |

PINN naturally inherits the advantages of conventional NN. PINN seamlessly integrates data with mathematical-physical models, even in partially understood, uncertain situations of imperfect data [38]. Due to the inherent smoothness or regularity of the PINN formulation, it is possible to find meaningful solutions even if the problem assumptions or models are incomplete [38]. PINN can directly handle nonlinear problems [27]. PINN is effective and efficient in dealing with the ill-posed and inverse problems, e.g., forward and inverse problems with no initial or boundary conditions specified, or problems where some parameters in the PDEs are unknown [16]. PINN does not need to deal with prohibitively small step sizes, so it can easily handle irregular and moving domain problems [12] and scales well in higher dimensions, but for PINN the sampling approach becomes more important [36,37].

It should be emphasized that PINN is not intended to replace ND such as FEM, but rather as a complement or alternative to ND methods in those suitable scenarios. PINN is particularly effective in solving, for example, the hypothetical and inverse problems; but for forward problems that do not require any data assimilation, existing ND solvers currently outperform PINN [12]. This simple yet powerful construction of PINN allows us to address a wide range of problems related to DEs, leading to the development of new data-efficient and physics-informed learning machines, new numerical solvers for DEs, and new data-driven methods for model inversion and system identification [16]. Thus, PINN is becoming a favorable catalyst for the emerging era of digital twins in various application fields [26,28,37,40,41].

Following the literature investigation, this study aims to further demonstrate the working mechanism and characteristic performance of soft-constrained PINN by solving simple yet general oscillator ODEs, especially the effectiveness and the robustness to noise of the PINN method with small or even minimalist data. Previously, Baty et al. [4] carried out a systematic and comprehensive benchmarking of PINN solving typical linear/nonlinear oscillator ODEs based on PyTorch. Ref. [2] addressed the problems of (i) limited known solution data, and (ii) integration intervals that, if too large, would make it difficult for the PINN to accurately predict the solution, when solving initial value problems (IVPs) for stiff ODEs. Improved strategies such as embedding more physical information, optimizing the training data loss to ensure that it considers all initial conditions more comprehensively, and incremental learning strategy, i.e., gradually increasing the integration intervals and optimizing the parameters of the PINN in each interval, and employing a shifting mesh to minimize the residuals of the DEs, have resulted in the improved PINN having a higher accuracy and a more stable training in solving IVPs for stiff ODEs process. In addition, these improved strategies have been shown to be equally effective in solving boundary value problems (BVPs), such as the solution to the high Reynolds number steady-state convection-diffusion equation. Ref. [3] further compared the performance of soft and hard constraints in solving higher-order Lane-Emden-Fowler type equations. The PINODE method [29] constructs reduced-order models (ROMs) based on an auto-encoder and aids model training with a physics-informed loss term. Using the collocation point technique, the residuals of the physics laws are added to the loss function as regularization terms, thus optimizing both data fitting and physical consistency during the training process.

Compared with these relevant literature on solving oscillator ODEs, our contributions mainly include the following folders:

- The characteristics and and applicable scenarios of PINN over conventional NN, traditional ND methods in solving DEs were surveyed and summarized through relevant literature investigation (see Table 1), enabling researchers especially engineers intrested in related fields to gain an overview quickly and easily.
- The mathematical framework and computational flow of the soft-constrained PINN is formalized suitable to solve both ODEs and PDEs. Through solving typical ODEs such as primer, Van der Pol and Duffing oscillators (covering first-order and second-order, linear and nonlinear), the working mechanism as well as performance characteristics of the soft-constrained PINN method are demonstrated, particularly on the effectiveness, efficiency comparable to Runge-Kutta, and robustness to noise of the PINN, with minimal data. The DeepXDE-based PINN implementation is light code and can be efficiently trained on both CPU and GPU platforms[2].
- The experimental verification comprehensively shows the excellent performance of the PINN method that: 1) PINN embeds physical information and prior knowledge (e.g., DEs and the law of conservation of energy) to essentially reduce data redundancy, thus has the unique advantage of achieving strong generalization with small data. It greatly reduces the need for labeled data and can even predict solutions with minimal data (i.e., first-order and second-order DEs require only one and two training data containing initial values, plus a few collocation points, respectively), no matter linear or nonlinear. 2) PINN is robust to noisy data and provides accurate and physically consistent predictions, with the aid of collocation points, can also extrapolates outside the time domain of the labeled training set, therefore the generalization ability is enhanced. 3) Training is accelerated when the gains obtained along with the reduction in the amount of data and fewer network parameters outweigh the delay caused by the increase in the loss function terms. 4) The soft-constrained PINN can easily impose a physical law (e.g., conservation of energy) constraint by adding a regularization term to the total loss function, improving the convergence performance of solving a second-order nonlinear Duffing oscillator with minimal data (2 training points plus some collocation points).

## 2 Data and Method

As mentioned above, PINN incorporates physical information and knowledge into network topology or computational processes as model priors, in a structured, modular neural network learning architecture. PINN back-propagates through the network and computes derivatives via automatic differentiation using chain rules, theoretically being able to accurately evaluate the differential operators at the collocation points with machine accuracy [4].

There are two specific implementations of PINN: the approaches of hard constraints and soft constraints. Hard constraints [20] generally ensure that physical knowledge or laws (e.g., differential equations, symmetries, conservation laws [21,29,33]), as well as boundary and initial conditions [17], are strictly adhered to by hard-coding or embedding them directly as part of the neural network architecture or computational process, which needs to be approximated by a specific network design or training strategy [35]. The soft constraints focused on here, on the other hand, are indirectly achieved by adding data and governing equation residual or regularization terms to the loss function, which is easier to implement and more flexible. This approach can be considered as a specific use case for multi-task learning. Soft constraints allow some flexibility or tolerance in satisfying physical laws or conditions, but require balancing predictive performance and physical accuracy by carefully adjusting the various weighting factors in the loss function.

In a nutshell, PINN is the process of improving the performance of learning algorithms with small data by utilizing a prior knowledge derived from our observations, experiences, and physical or mathematical understanding of the world [16], which prompts reducing data redundancy, improving generalization, increasing computational efficiency, and also increasing robustness and interpretability.

[2] The experimental codes are available at: https://github.com/mikelu-shanghai/PINNtoODEwithSmallData .

### *2.1 Soft-constrained PINN*

Consider a physical system defined on a spatial or spatio-temporal domain $\Omega \subseteq \mathbb{R}^d$, where the unknown $u(\boldsymbol{x}): \mathbb{R}^d \to \mathbb{R}^m$ is system state variable that are functions of spatial or temporal coordinates $\boldsymbol{x} \in \Omega$. For time-independent systems, $\boldsymbol{x} = (x_1, ..., x_d)$; for time-dependent systems, $\boldsymbol{x} = (x_1, ..., x_{d-1}, t)$. The physical laws of the dominant system are often characterized in terms of ODEs/PDEs, and these equations are known as the governing equations [13], given by

Differential equation: $\mathcal{F}(u; \theta)(\boldsymbol{x}) \equiv \mathcal{F}(u, \boldsymbol{x}; \theta) = 0, x \in \Omega.$ (1)

Initial conditions: $\mathcal{I}(u; \theta)(\boldsymbol{x}, t_0) = 0, x \in \Omega_0.$ (2)

Boundary conditions: $\mathcal{B}(u; \theta)(\boldsymbol{x}, t) = 0, x \in \partial\Omega.$ (3)

For time-dependent systems (i.e., dynamical systems), initial conditions Eq. (2) need to be set for the state variables (and sometimes their derivatives) at the initial moment $t_0$. $\theta \in \Theta$ parameterizes or controls the system, where $\theta$ can be a vector or a function included in the control equation. For systems characterized by PDEs, it also needs to constrain the state variables on the boundaries of the spatial domain $\partial\Omega$ in order to make the system well-posed. For the boundary points $x \in \partial\Omega$, we have boundary conditions Eq. (3). If there are no constraints on the initial and boundary conditions, then $\mathcal{I}(u; \theta) \triangleq 0$ and $\mathcal{B}(u; \theta) \triangleq 0$ [13].

As shown in Fig. 1, suppose there is a system obeying Eq. (1) and a dataset $\{u(\boldsymbol{x}_i)\}_{i=1,...N}$. It is then possible to construct the neural network $u_w(\boldsymbol{x})$ and train it with the following loss function, namely

$$\begin{aligned}\mathcal{L}_{total} &= \mathcal{L}_{data} + \underbrace{\mathcal{L}_{gov} + \mathcal{L}_{IC} + \mathcal{L}_{BC} + \mathcal{L}_{reg}}_{\mathcal{L}_{phys}} \\ &= \frac{\lambda_d}{N}\sum_{i=1}^{N} \| u_w(\boldsymbol{x}_i) - u(\boldsymbol{x}_i)\|^2 + \frac{\lambda_g}{|\Omega|}\int_{\Omega} \| \mathcal{F}(u_w; \theta)(\boldsymbol{x})\|^2 \mathrm{d}\boldsymbol{x} \\ &+ \frac{\lambda_i}{|\Omega_0|}\int_{\Omega_0} \| \mathcal{I}(u_w; \theta)(\boldsymbol{x})\|^2 \mathrm{d}\boldsymbol{x} + \frac{\lambda_b}{|\partial\Omega|}\int_{\partial\Omega} \| \mathcal{B}(u_w; \theta)(\boldsymbol{x})\|^2 \mathrm{d}\boldsymbol{x} + \mathcal{L}_{reg}.\end{aligned} \quad (4)$$

Where, $\mathcal{L}_{data} = \frac{1}{N}\sum_{i=1}^{N} \| u_w(\boldsymbol{x}_i) - u(\boldsymbol{x}_i)\|^2$ is the regular data loss of PINN in matching the labeled datasets; $\mathcal{L}_{gov} = \int_{\Omega} \| \mathcal{F}(u_w; \theta)(\boldsymbol{x})\|^2 \mathrm{d}\boldsymbol{x}$ is the residual loss that makes the network satisfy the governing equation constraints; $\mathcal{L}_{IC} = \int_{\Omega_0} \| \mathcal{I}(u_w; \theta)(\boldsymbol{x})\|^2 \mathrm{d}\boldsymbol{x}$ and $\mathcal{L}_{BC} = \int_{\partial\Omega} \| \mathcal{B}(u_w; \theta)(\boldsymbol{x})\|^2 \mathrm{d}\boldsymbol{x}$ are the corresponding losses that make PINN satisfy the initial and boundary conditions, respectively.

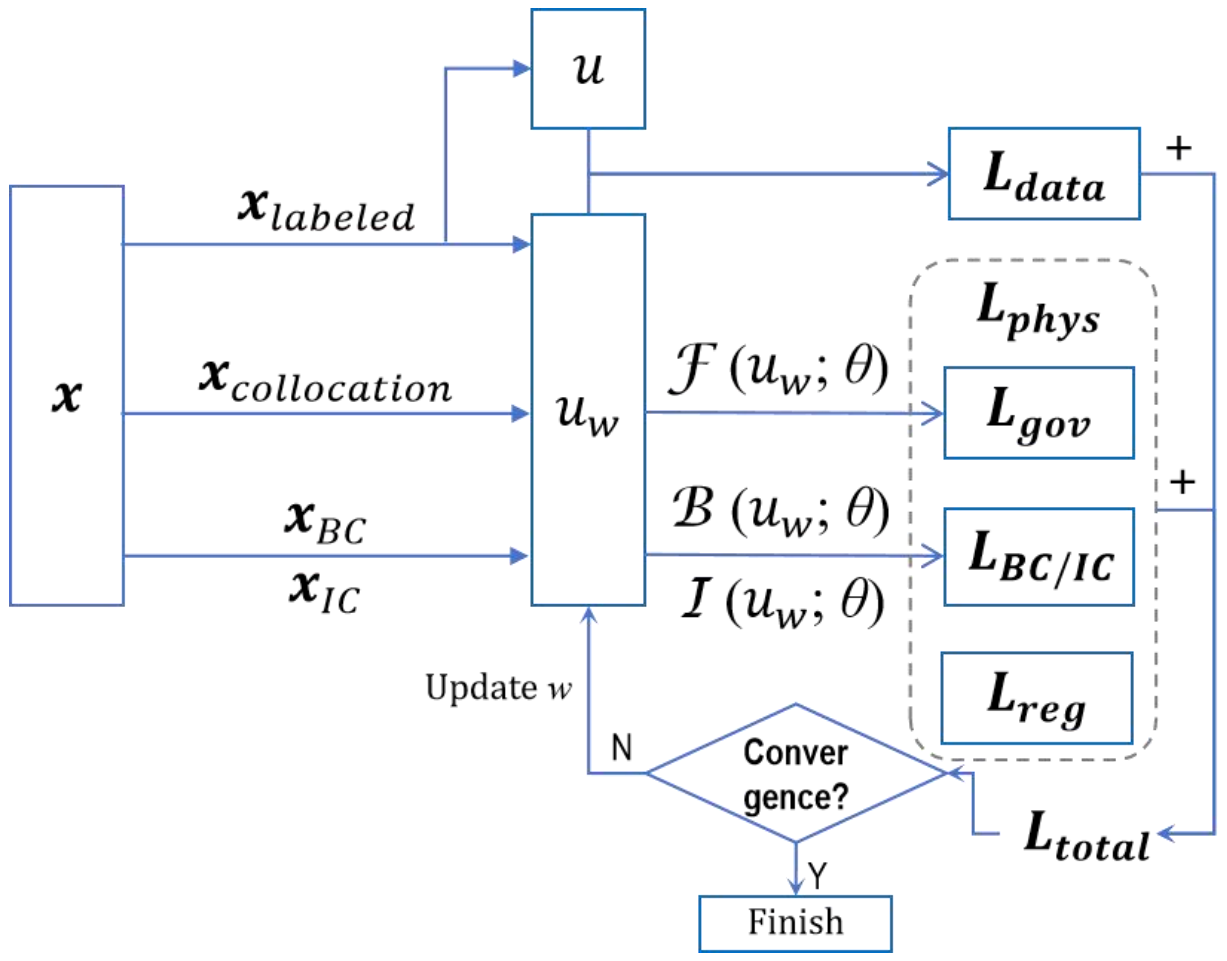

**Figure 1:** Schematic diagram of physics-informed neural network via soft constrains

PINN losses are flexible and scalable: for example, the extra optional in Eq. (4) can be a regularization term $\mathcal{L}_{reg}$ that satisfies the law of conservation of energy; If there are no available data or initial/boundary constraints, the corresponding loss terms can simply be omitted. The learning weights of these losses can be set by adjusting the hyper-parameters $\lambda_d, \lambda_g, \lambda_i, \lambda_b, \lambda_r$.

In order to compute (4), several integral terms need to be evaluated, which involve the computation of higher-order derivatives of $u_w(\boldsymbol{x})$. PINN utilizes automatic differentiation of computational maps to calculate these derivative terms. The integrals are then approximated using a set of collocation points, which can be sampled using the Monte-Carlo method. $N_d, N_g, N_i, N_b$ is used to denote the number of corresponding data or collocation points, and $\mathcal{D}_d, \mathcal{D}_g, \mathcal{D}_i, \mathcal{D}_b$ to denote corresponding data sets. Then, the loss function can be approximated as [13]

$$\mathcal{L}_{total} = \frac{\lambda_d}{N_d}\sum_{i=1}^{N} \| u_w(\boldsymbol{x}_i) - u(\boldsymbol{x}_i)\|^2 + \frac{\lambda_g}{N_g}\sum_{i=1}^{N_g} \| \mathcal{F}(u_w;\theta)(\boldsymbol{x}_i)\|^2 + \frac{\lambda_i}{N_i}\sum_{i=1}^{N_i} \| \mathcal{I}(u_w;\theta)(\boldsymbol{x}_i)\|^2 + \frac{\lambda_b}{N_b}\sum_{i=1}^{N_b} \| \mathcal{B}(u_w;\theta)(\boldsymbol{x}_i)\|^2 + \mathcal{L}_{reg}. \quad (5)$$

Eq. (5) can be efficiently trained using first-order optimizer such as SGD and second-order L-BFGS.

### *2.2 Data and Implementation*

The soft-constrained PINN is implemented via DeepXDE [19] and compared the results with a PyTorch-based implementation [4], so the same linear and nonlinear oscillator ODEs , i.e., tutorial example and Van der Pol oscillators were chosen to solve. The main focus is to study the efficiency and accuracy performance characteristics of the PINN method in solving DEs in small or even minimalist data scenarios. Therefore, for the sake of simplicity, typical oscillator ODEs (covering first-/second-order, linear/nonlinear) were chosen to facilitate intuitive comparisons. As case study experiment, the numerical solution obtained by the Runge-Kutta method is considered to be ground truth. In practice, however, the ground truth can often be given by experimental or field measurements, and therefore it is not necessary to rely on ND methods such as Runge-Kutta. The generation of training points, collocation points and noise in the experiments will be detailed in the corresponding sections of “3 Results and Discussion”. It should be noted that the PINN method formalized here is also applicable to other types of DEs such as PDEs.

DeepXDE is well-structured and highly configurable. Code written in DeepXDE is shorter, closer to mathematical formulas, and more computationally efficient. Solving DEs in DeepXDE is akin to "building blocks" using built-in modules that specify the computational domain (geometric and temporal), differential equations, boundary/initial conditions, constraints, training datasets, neural network structure, optimization algorithms, and hyper-parameters, etc., all of which are loosely coupled. The codes for primer, Van der Pol and Duffing oscillators are open-sourced at: https://github.com/mikelu-shanghai/PINNtoODEwithSmallData .

## 3 Results and Discussion

### *3.1 PINN v.s. NN to Solving Linear Oscillator ODE*

The linear oscillator ODE (named the primer oscillator) is

$$\frac{du}{dt} + 0.1u - \sin{(\pi t/2)} = 0. \tag{6}$$

Where, $t \in [0{,}30]$, initial condition $u_0 = 1$. We solved it with conventional NN and PINN respectively, and both used a 3-layer 32-neuron (3×32) fully connected hidden layer as base network, Adam optimizer, learning rate $\eta = 3 \times 10^{-3}$. The numerical solution obtained by the Runge-Kutta method (3000 steps, at least about 200 steps required for stability and accuracy) is regarded as ground truth (or exact solution).

When solving by NN, if the number of training points is insufficient or not well distributed, for example: 1) Case 1 in Fig. 2a samples 26 training points uniformly in the entire time domain but does not converge well; 2) Case 2 in Fig. 3a samples 61 training points uniformly in the left half of the time domain. Although the left half of the interval is well fitted, it may fail completely on the right half of the interval without training points, i.e., there is no extrapolation capability. In this case, about 50 or more training points need to be sampled uniformly over the entire time domain to get a good match with the numerical solution (regarded as exact solution). For these oscillators without noise interference, solving via Runge-Kutta usually takes within 1 second.

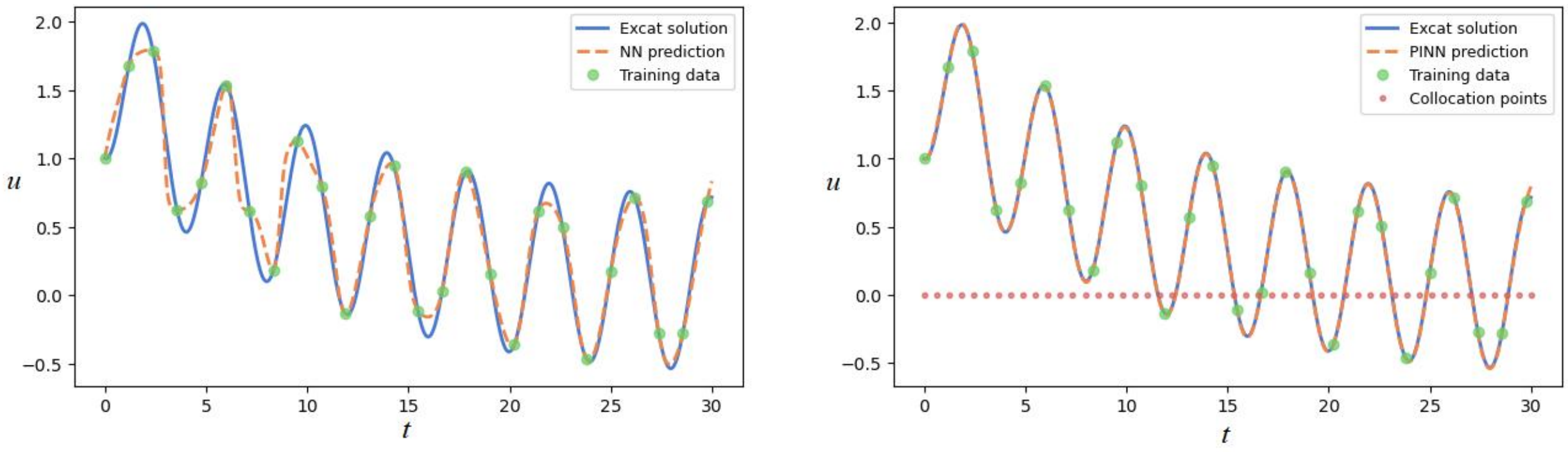

(**a**) No. of training data $N_{data} = 26$, No. of collocation points $N_c = 0$

(**b**) $N_{data} = 26, N_c = 50$

**Figure 2:** Result comparison of PINN v.s. NN solving linear oscillator – Case 1. (Reproduced from [4] via DeepXDE.)

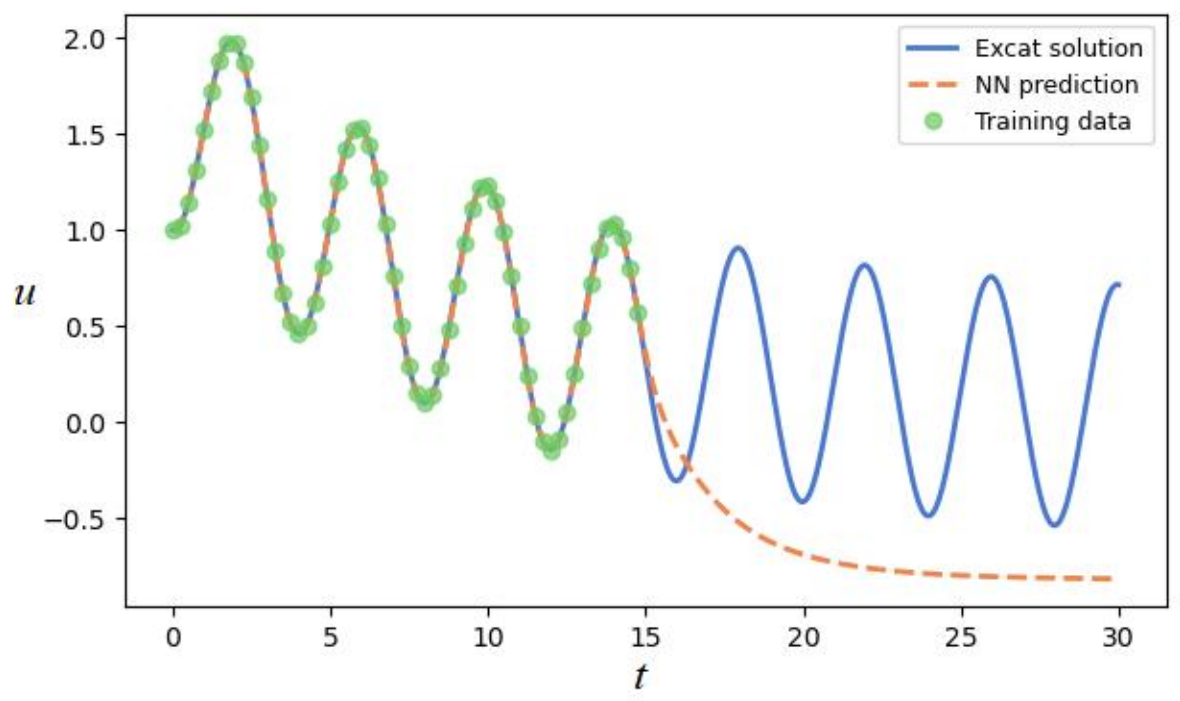

(**a**) $N_{data} = 60$ (Only left interval), $N_c = 0$

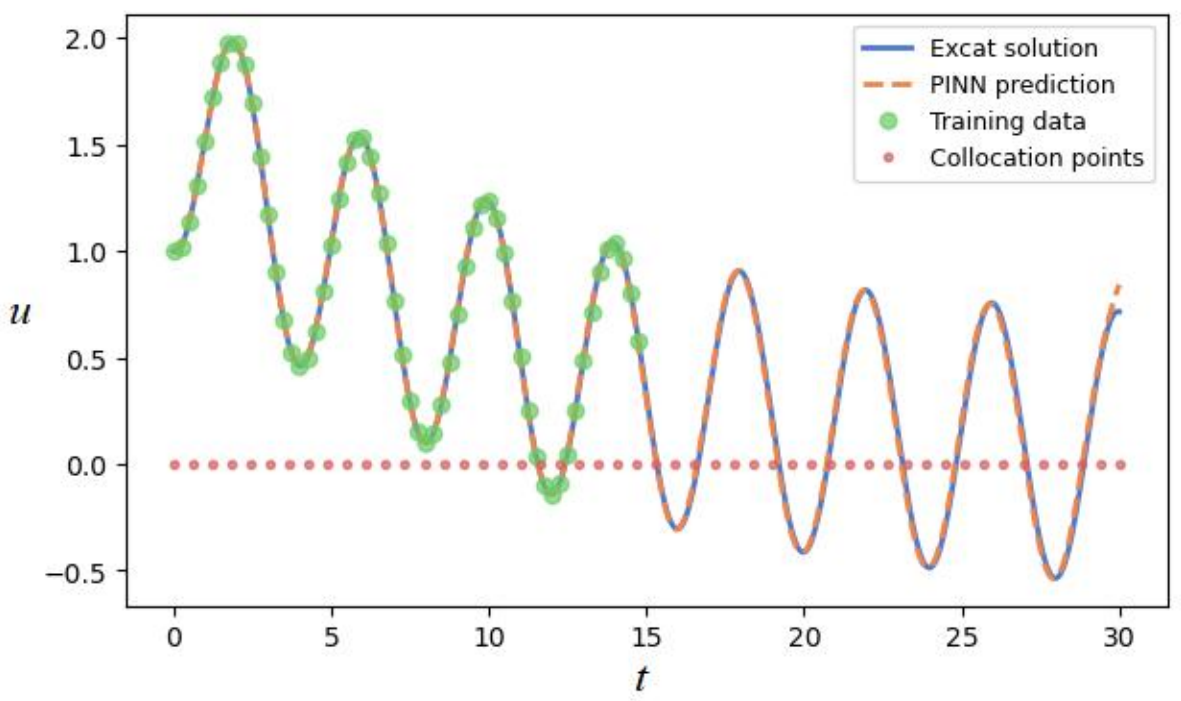

(**b**) $N_{data} = 60$ (Only left interval), $N_c = 50$

**Figure 3:** Result comparison of PINN v.s. NN solving linear oscillator – Case 2. (Reproduced from [4] via DeepXDE.)

The only difference between solving by the soft-constrained PINN approach in this example and a conventional NN is the addition of $\mathcal{L}_{phys}$ to the total loss function. Since the initial condition in the conventional NN was already included in the training set (i.e., the first training point on the left), only $\mathcal{L}_{gov}$ satisfying Eq. (6) was added to total loss. This is accomplished by adding extra collocation points (see Eq. (5) ), in this example sampling at least 50 collocation points uniformly throughout the entire time domain. The exact number of collocation points required depends on the number of neural network parameters, i.e. the complexity of the problem being solved. The distribution of collocation points (uniform or random) also has an impact on the results [36]. The learning rate and loss weights also affect the convergence of the gradient descent algorithm [4]. Here the loss weights were taken as $\lambda_d = 1.0, \lambda_r = 6 \times 10^{-2}$ through hyper-parameter optimization. The loss weights act as a balance between the dual drivers of data and physical information.

For the two cases where NN fails to converge, the results obtained by the PINN method are shown in Fig. 2b and Fig. 3b, respectively, and a significant improvement can be seen when comparing with the corresponding sub-figure (a) on the left: 1) The number of labeled training data points required is reduced to 26 (the minimalist case in this example actually requires only the 1 initial value point plus another 48 and more unlabeled collocation points, see in Fig. 4). Thus, the need for sampling points especially labeled data is drastically (even orders of magnitude) reduced by the PINN method - one of its significant advantages. 2) In particular, the generalization ability is enhanced by the assistance of collocation points and the use of physical information carried by ODEs, which equips PINN with the ability to extrapolate data outside the time domain of the training set. In addition, once PINN has completed its training, the prediction for new data sample is instantaneous, which is a property that traditional ND methods (e.g., Runge-Kutta) do not have.

### *3.1.1 Minimalistic Training Data*

A natural question here is that how minimally dependent is PINN on data especially on labeled data? For this example, if only one point, the initial value, is taken and the collocation points are kept to be 48, a good solution is obtained, as shown in Fig. 4a; Simply increasing the collocation points (e.g., to 80) yields a better result as shown in Fig. 4b. It has already been showed that when the non-linearity of the problem is weak, a very small amount of labeled training data plus a few collocation points is sufficient to predict the solution. In the minimalist case, just as classical analytic or ND methods require only one and two initial conditions for solving first-order and second-order differential equations, respectively [4]. We also find out that PINN require only one training point (initial value) and two training points containing the initial value point, respectively, no matter linear or nonlinear oscillator ODE (refer to the result of second-order nonlinear Duffing oscillator in Fig. 7b).

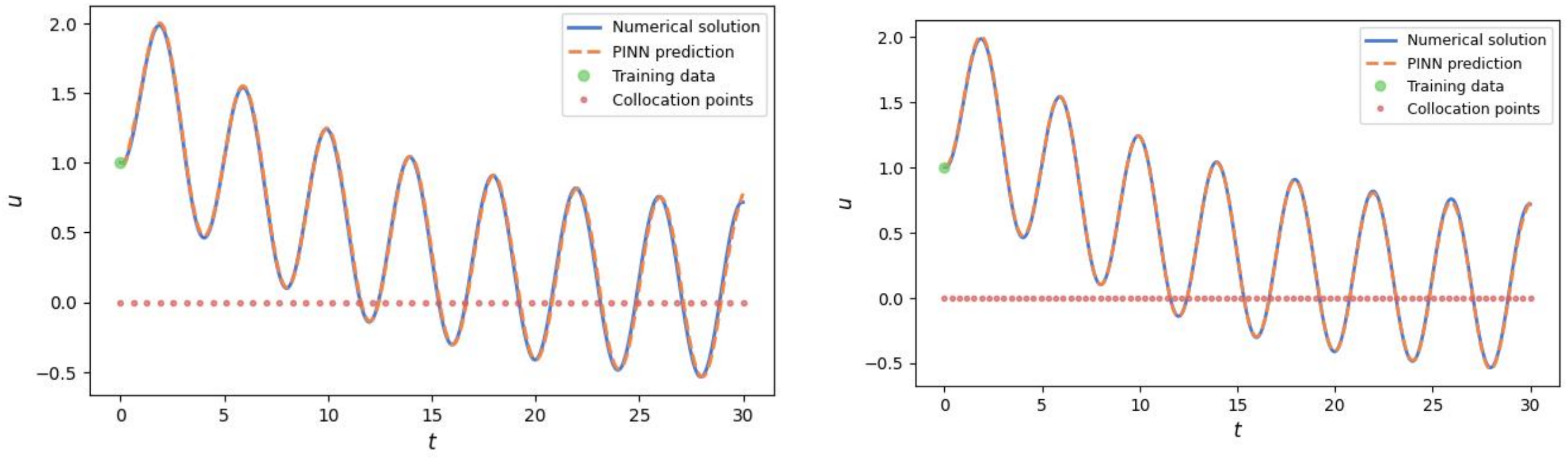

(**a**) $N_{data} = 1,\ N_c = 48$ (**b**) $N_{data} = 1,\ N_c = 80$

**Figure 4:** Minimalistic training data example of first-order linear ODE. (Reproduced from [4] via DeepXDE.)

### *3.1.2 Influence on Training Time*

On one hand, compared with NN, PINN can significantly shorten the training time because it requires only small data (minimal labeled training data plus a small number of collocation points) and fewer parameters of network. On the other hand, as its loss function becomes more complex with the addition of $\mathcal{L}_{phys}$, it requires more computations and sometimes even more iteration epochs when training, which in turn drags down the training efficiency.

A validation test of primer oscillator is performed via DeepXDE 1.11.1 based on a CPU laptop (Windows10, Intel Core i9-9900K @3.6GHz, 32GB Memory). The training time is the average of 5 training sessions (24,000 epochs per session), usually, 24,000 epochs are not needed before convergence thus the training time is conservative, and 24,000 epochs are trained for fair comparison between different settings, and the results are shown in Fig. 5. It can be seen that, for linear oscillator ODEs, the training is accelerated when the gains obtained along with the reduction in the amount of data outweigh the delay caused by the increase in the loss function terms. Note that, for ease of comparison, the same base network (3×32) is used. PINN can accelerate the training even further by reducing the number of network parameters, because PINN need less network parameters than NN due to its small data feature. For example, for Fig. 5b, if the base network is shrunk to 3×16, it also converges well while the training time reduced to 14.30s. In fact, about 7,000 epochs is sufficient for convergence in solving this primer oscillator, even less epochs if using early-stopping technique. So the training time has the potential to be reduced to 3 seconds or even less.

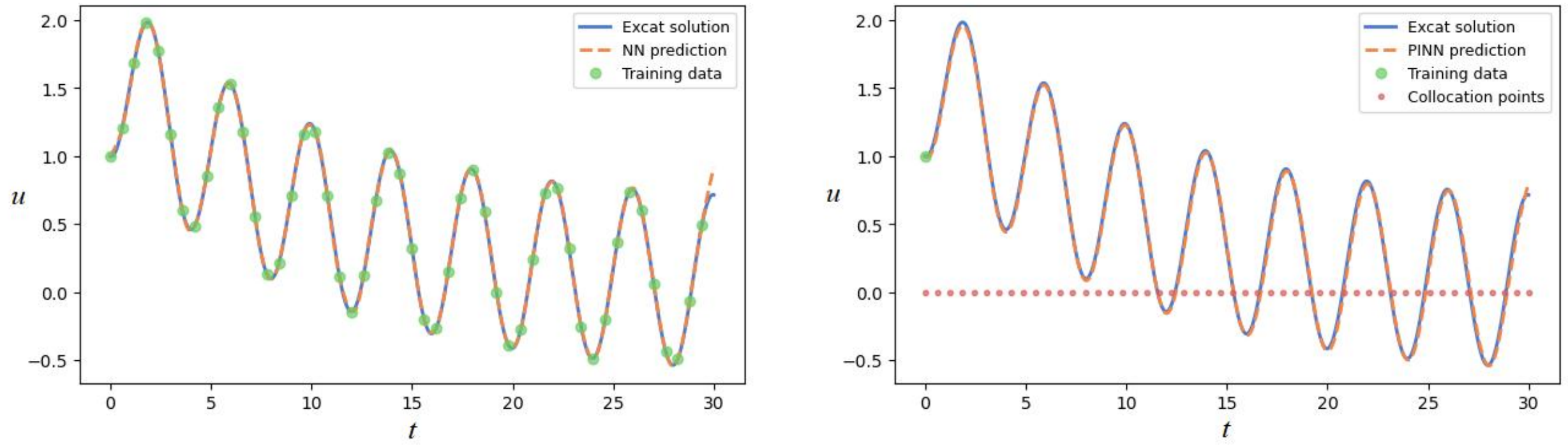

(**a**) Training time: 16.56s; $N_{data} = 50,\ N_c = 0$ (**b**) Training time: 15.23s; $N_{data} = 1,\ N_c = 48$

**Figure 5:** Training time of PINN v.s. NN solving primer oscillator via DeepXDE.

### *3.2 PINN Solving Nonlinear Oscillator ODE*

The soft-constrained PINN also performs well in solving nonlinear oscillator ODEs. The typical strongly nonlinear Van der Pol oscillator is successfully solved via DeepXDE (backend: tensorflow.compat.v1), further demonstrating the tolerance of PINN to noisy data as well as the light-code and fast training speed of the DeepXDE implementation of PINN. The ODE of Van der Pol is

$$\frac{d^2u}{dt^2} + \omega_0^2 u - \epsilon\,\omega_0(1-u^2)\frac{du}{dt} = 0. \tag{7}$$

Where, $t \in [0,1.5]$, initial condition $u_0 = 1$; $\omega_0$ is the normalized angular velocity, taken as $\omega_0 = 15$; $\epsilon$ reflects the degree of non-linearity of Van der Pol, i.e. the larger the value of $\epsilon$ the stronger the non-linearity.

#### *3.2.1 Noise Tolerance of PINN with Small Data*

The nonlinear Van der Pol oscillator is solved with small and noisy data. The same as linear primer oscillator, the fully connected base network of 3×32 was still used, which reflects the strong expressive power of neural networks inherited by PINN. The relevant parameter settings (i.e., $\epsilon$, number of training data points, number of collocation points, normal noise variance) are detailed in Fig. 6. The solution by numerical method is considered to be exact solution or ground truth. It can be seen from Fig. 6: 1) With the same amount of training data and collocation points, the results obtained from training with noisy data (right column) are basically identical to the noiseless results (left column), showing the good inclusiveness of the PINN method against noise. 2) As the degree of non-linearity increases (by increasing $\epsilon$), more training and collocation points are needed to be arranged in regions with significant nonlinear features to obtain an exact solution. That is, when $\epsilon = 1, 3, 5$, the least number of training and collocation points are $N_{data} = 28, 32, 38$; $N_c = 15, 25, 40$, respectively. And 3) as the non-linearity increases, the tolerance to noise decreases: when $\epsilon$ increases from 1 to 5, the $\sigma$ that can be tolerated decreases from 0.1 to 0.05; even increasing the number of data points has a limited effect on improving the ability to tolerate noise. This may be due to the fact that both non-linearity and noise introduce computational complexity, while PINN cannot accommodate both.

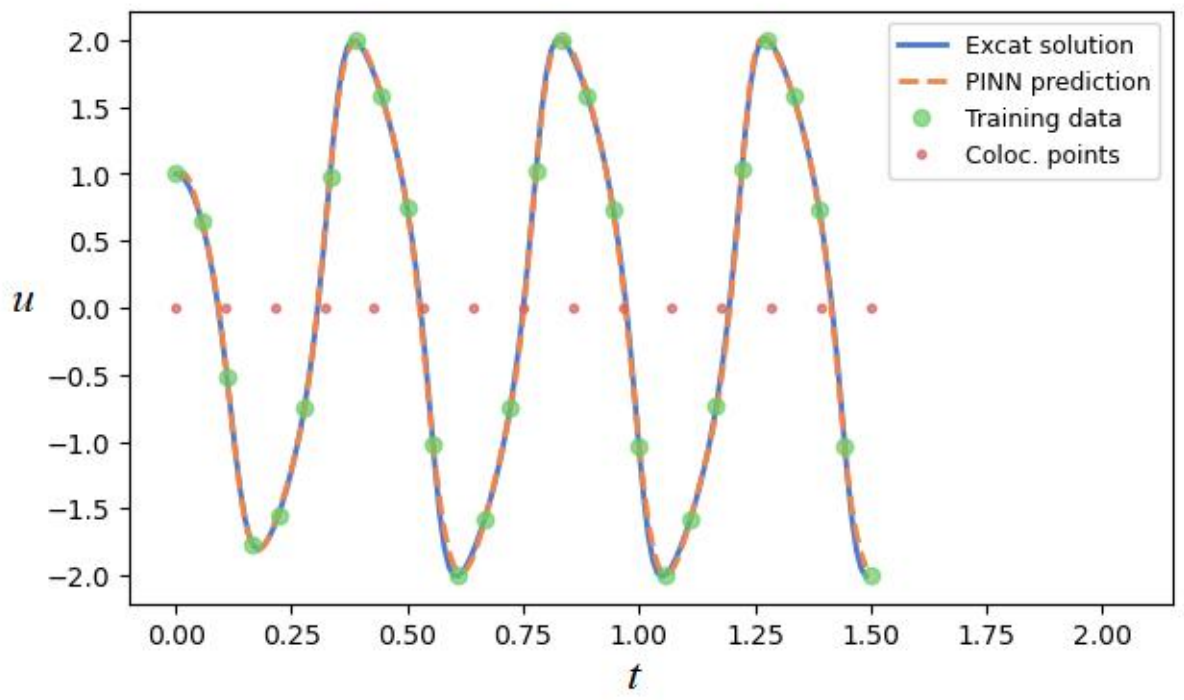

(**a**) $\sigma = 0$; $N_{data} = 28, N_c = 15$; $\epsilon = 1$

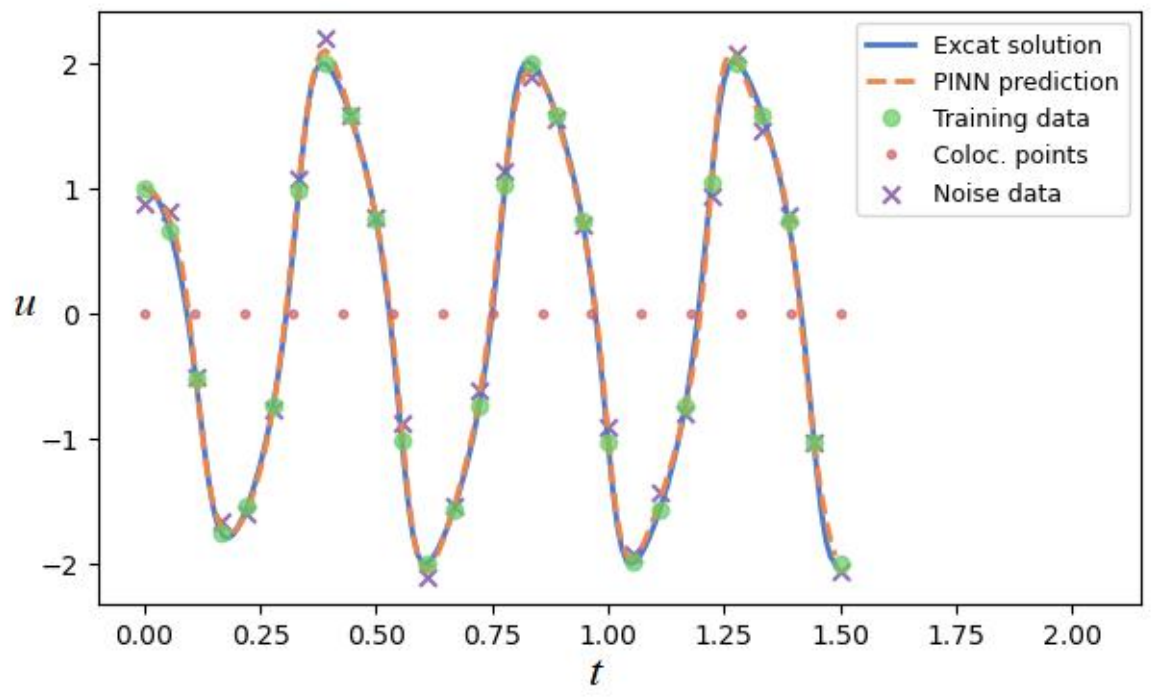

(**b**) $\sigma = 0.1$; $N_{data} = 28, N_c = 15$; $\epsilon = 1$

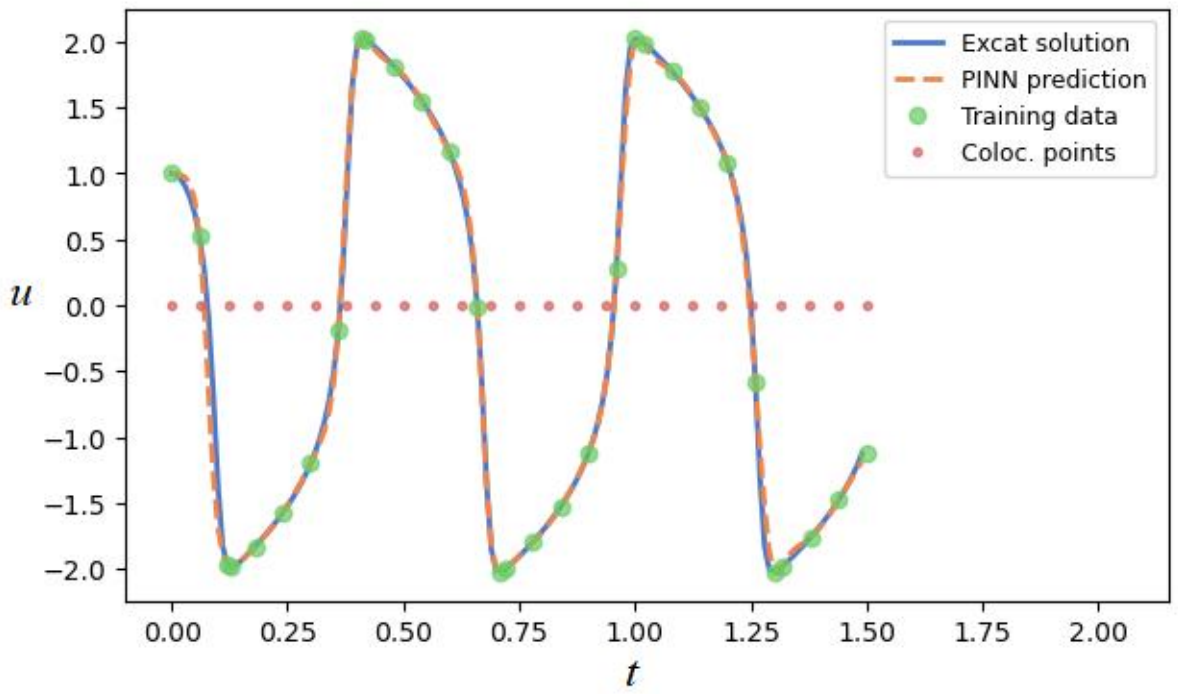

(**c**) $\sigma = 0;\ N_{data} = 32, N_c = 25;\ \epsilon = 3$

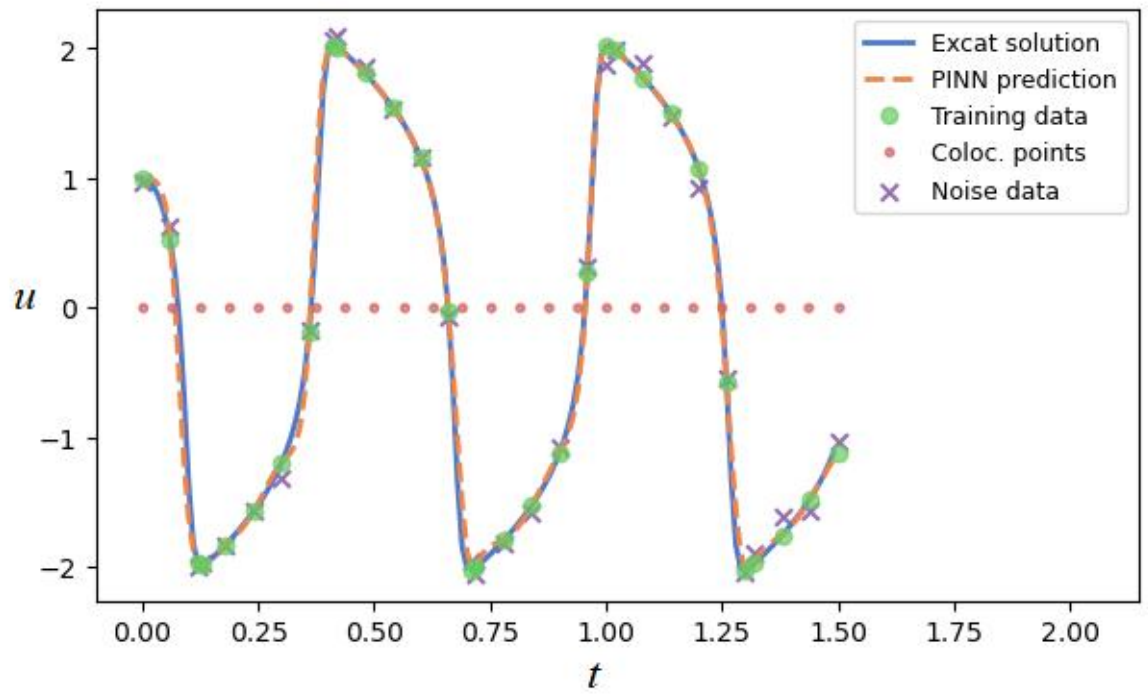

(**d**) $\sigma = 0.08;\ N_{data} = 32, N_c = 25;\ \epsilon = 3$

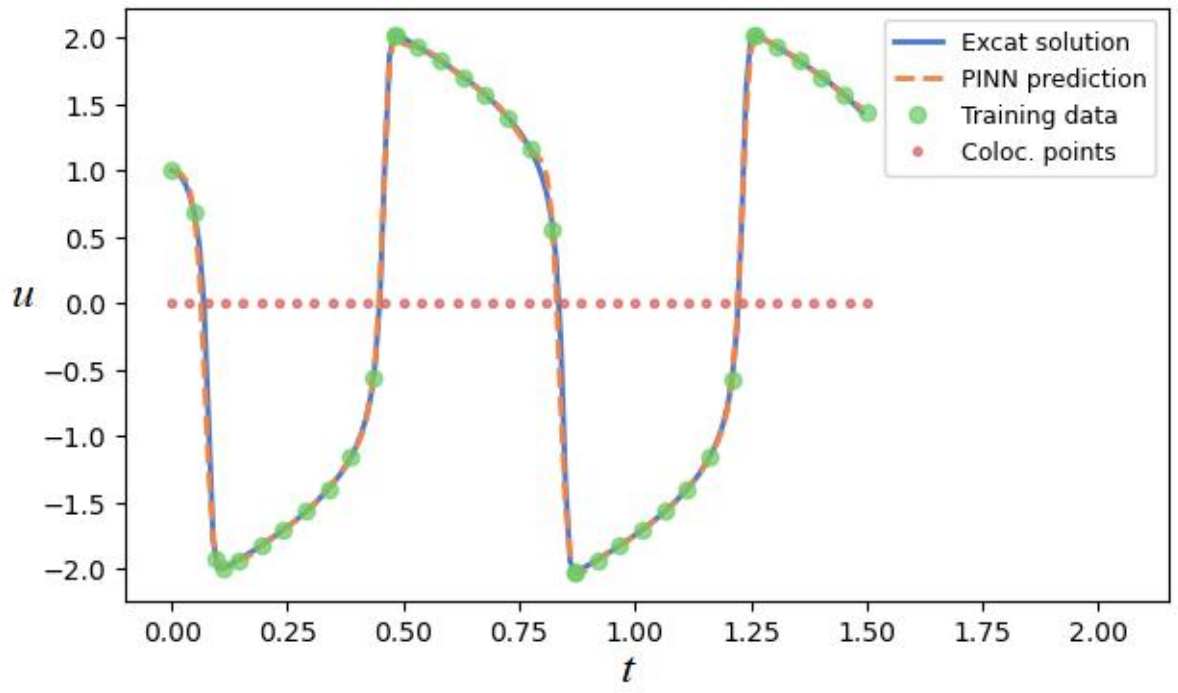

(**e**) $\sigma = 0;\ N_{data} = 38, N_c = 40;\ \epsilon = 5$

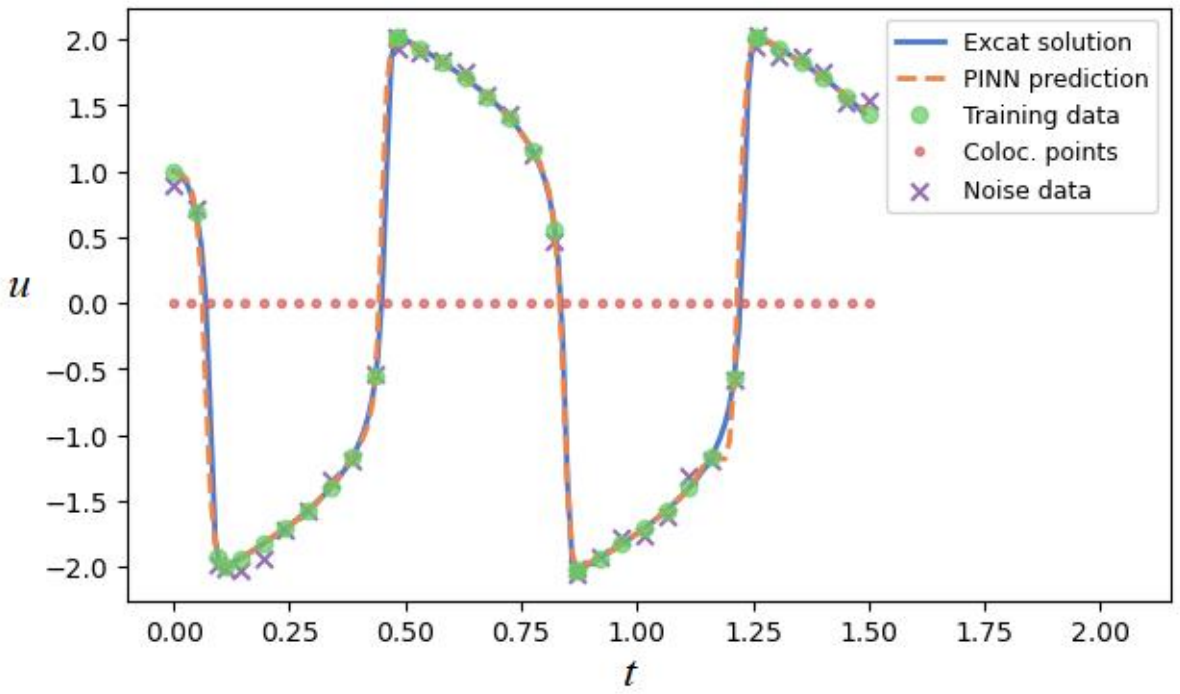

(**f**) $\sigma = 0.05;\ N_{data} = 38, N_c = 40;\ \epsilon = 5$

**Figure 6:** Results of PINN solving Van der Pol oscillator via DeepXDE with noisy and small data.

### *3.2.2 Implementation Code via DeepXDE v.s. PyTorch*

Comparison of code Listing 1 and Listing 2 in Appendix A shows that the overall process (seen in Fig. 1) of realizing soft-constrained PINN based on DeepXDE or PyTorch is consistent. DeepXDE has a better customized encapsulation of the PINN method and thus requires less coding. With the same setup for solving Van der Pol, the number of code lines (about 40 lines) via DeepXDE implementation is roughly one-third of the PyTorch implementation. In addition, the training efficiency on PINN of DeepXDE is significantly better than that of PyTorch, thanks to the backend (tensorflow.compat.v1) [19]. For the example in Figure 5b, PyTorch-based training took 35.6 s with the same settings, which is about 2 times more than that of DeepXDE.

### *3.2.3 Computational Efficiency @CPU v.s. @GPU*

PINN does not depend on big data so it can be efficiently computed on platforms such as CPU or GPU. The training duration for solving primer, Van der Pol oscillators via DeepXDE are compared on CPU and GPU (CPU as before, GPU is an NVIDIA GTX1080Ti) respectively. Take the average of 5 training sessions (24,000 epochs each session), and base networks use 2×32, 3×32 fully connected hidden layers, respectively. As shown in Table 2, the training acceleration effect of GPU on PINN solving primer and Van der Pol is obvious: the training duration is reduced from 16.02s, 17.60s to 8.73s, 13.15s, respectively. When the base network is shrunk to 3×16, the training duration results show a consistent trend. In fact, about 8,000 epochs of training is sufficient for convergence in this example. So the training duration in Table 2 is conservative, and there is still a large margin reduction to about one-third.

**Table 2:** PINN training duration (second) @CPU v.s. @GPU via DeepXDE

| PINN to ODEs via DeepXDE | @CPU | @GPU |
|---|---|---|
| Primer Oscillator ($N_{data} = 50, N_c = 48$) | 16.02 | 8.73 |
| Van der Pol Oscillator ($N_{data} = 50, N_c = 48$) | 17.60 | 13.15 |

### *3.3 Conservation of Energy Regularization Improves Result of Solving Duffing Oscillator*

A major advantage of PINN is that it can fully utilize physical information or knowledge as a prior (e.g., the law of conservation of energy). In the soft-constrained approach, it is implemented by adding the regularization term $\mathcal{L}_{reg}$ to the total loss function. Duffing is a second-order nonlinear oscillator

$$\frac{d^2u}{dt^2} + \alpha u + \beta u^3 = 0. \tag{8}$$

Where, $t \in [0,1.5]$, take $\alpha = 1.0, \beta = 1.0$ with initial condition $u_0 = 15, u'_0 = 0$. Again, the exact solution was obtained by the Runge-Kutta method. The conservation of energy term of Eq. (8) is $E = \frac{1}{2}(\frac{du}{dt})^2 + \frac{1}{2}\alpha u^2 + \frac{1}{4}\beta u^4$, substituting to the regularization term $\mathcal{L}_{reg}$.

Considering the minimalistic data situation, for second-order ODEs only two training points containing initial values were used, and the number of collocation points $N_c = 40$, chosen uniformly. The optimal tuning results of the PINN solving Duffing oscillator with and without the conservation of energy regularization term are shown in Fig. 7a,b, respectively. The PINN without the conservation of energy regularization in the minimalistic data case, which is affected by non-linearity, fails to complete convergence throughout the entire 72,000 epochs of training. In contrast, the PINN with the conservation of energy regularization achieves a significant improvement, which is in perfect agreement with the exact solution. It should be noted that the conservation of energy regularization is useful for Duffing as it follows the conservation of energy; It is useless for Van der Pol because Van der Pol is energy dissipative.

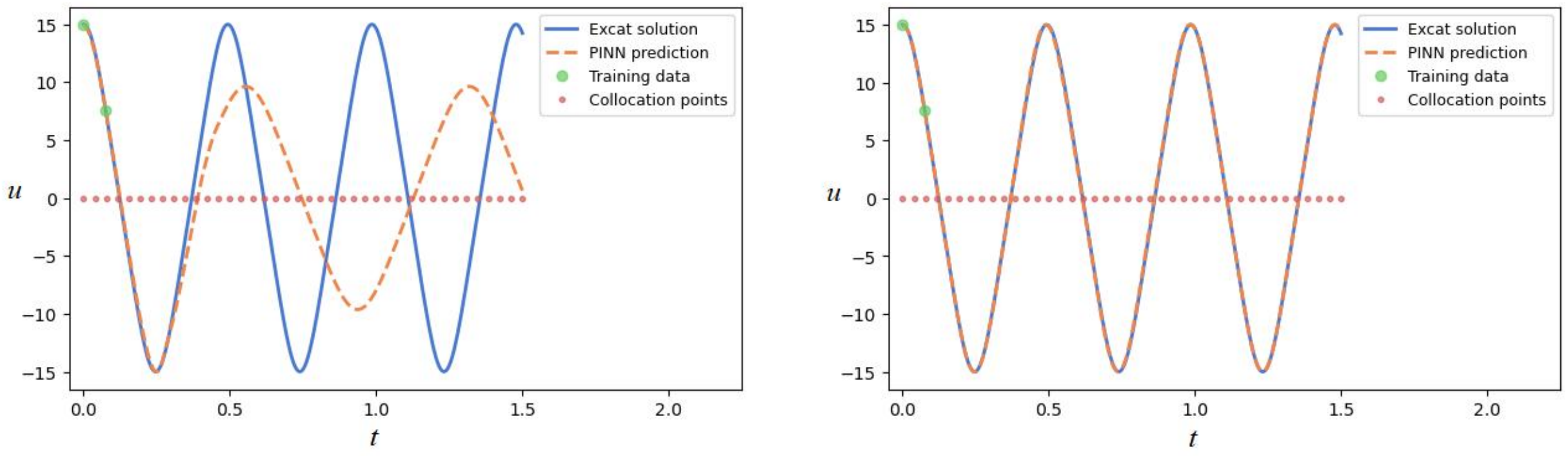

(**a**) w/o the conservation of energy regularization (**b**) with the conservation of energy regularization

**Figure 7:** Results of PINN solving Duffing oscillator with the conservation of energy regularization or not ($N_{data} = 2, N_c = 40$)

## 4 Conclusions

The success of deep neural networks relies on big data and strong computational power, yet neglects to fully utilize prior information or knowledge, thus it lacks robustness or even fails to converge in small data scenarios. Physics-informed neural network (PINN) inherits the strong expressive ability of neural networks (NN) and incorporates physical information and knowledge into neural network topology architecture as well as computational process by means of hard-coding or soft constraints, which reduces data redundancy, improves generalization performance, and consequently increases computational efficiency, robustness, and interpretability. The features and applicable problems of PINN over conventional NN, traditional numerical discretization (ND) methods (e.g., Runge-Kutta, FEM) for solving differential equations (DEs) were distilled and summarized in Table 1 through a mini literature survey. While there are challenges for vanilla Runge-Kutta to integrate noise seamlessly, PINN has the unique advantage of achieving strong generalization with small and noisy data.

The mathematical framework and computational flow of soft-constrained PINN focused in this study for solving ODEs/PDEs is formulated. It is clear from the composition of the loss function that PINN is driven by both data and DEs or physical information. Through solving simple yet general ODEs such as primer, Van der Pol and Duffing oscillators (covering first-order and second-order, linear and nonlinear), the performance characteristics (especially the precision, efficiency and robustness to noise) of the soft-constrained PINN method with minimal data were examined. The experimental results show that:

1) The soft-constrained PINN greatly reduces the need for labeled data. A very small amount of labeled training data plus a few collocation points no need to be labeled are sufficient to predict the solution of oscillator ODEs as precise as Runge-Kutta. In the minimalist case, only one or two training points (including initial values) are needed for first-order or second-order ODEs respectively, just as classical analytic or ND methods. Even strongly nonlinear oscillators such as Van der Pol require only an appropriate increase in the number of training or collocation points.
2) Small data allows fewer network parameters, which improves computational efficiency. The training duration of PINN, usually in seconds for the scalar oscillator ODEs, is accelerated when the gains obtained along with the reduction in the amount of data outweigh the delay caused by the increase in the loss function terms. It has the potential to approach to the computation time of Runge-Kutta. In addition, the DeepXDE-based implementation of PINN is not only light code but also efficient training on both GPU and CPU platforms.
3) With the aid of the embedded DEs or physical knowledge, PINN has the ability to extrapolate data outside the time domain of the training set, and is robust to noisy data, thus with enhanced generalization capabilities and better interpretability. Solving the second-order nonlinear Duffing oscillator with regularization by conservation of energy is able to converge quickly with minimal data (two training points plus some collocation points), whereas it does not converge without the regularization, indicating that the soft-constrained PINN can conveniently impose a physical law (e.g., the law of conservation of energy) constraint by adding a regularization term to the total loss function, thus improving the performance to deal with various complexities such as nonlinearity.

Due to the excellent characteristic performance, PINN is becoming a favorable catalyst for the era of Digital Twins, in the fields where the mathematical equations are not yet clear or incomplete.

However, there are some issues still worth further exploration, and can be our future work: (1) In this study, the time domain results were mainly analyzed and discussed. The accuracy of PINN in frequency (especially high-frequency) response results warrants further investigation. (2) Whether the conclusions drawn from typical ODEs such as primer, Van der Pol and Duffing oscillators (simple without loss of generality, covering linear/nonlinear, first-/second-order) as case study, can be directly extrapolated to other nonlinear complex cases or not, worth further research. Theoretical or quantitative scalability influenced by system complexity or dimensionality can also be studied specifically, particularly in conjunction with application scenarios. (3) This study primarily applied Gaussian additive noise to the Van der Pol oscillator, and applying the PINN methods to stochastic differential equations can also be a promising direction.

**Acknowledgement:** The author would like to thank Professor Hubert Baty from University of Strasbourg for his inspiration, and acknowledges the valuable suggestions from the peer reviewers.

**Funding Statement:** This research is supported by ongoing institutional funding (BBYQ202415) and sponsored by Ministry Education Project of China (20230104148).

**Author Contributions:** The author confirm contribution to the paper as follows: Conceptualization, methodology, writing, software, validation, funding acquisition, Kai-liang Lu.

**Availability of Data and Materials:** The data and code that support the findings of this study are openly available in Github at https://github.com/mikelu-shanghai/PINNtoODEwithSmallData.

**Ethics Approval:** Not applicable.

**Conflicts of Interest:** The authors declare no conflicts of interest to report regarding the present study.

## Nomenclature

| | |
|---|---|
| $u$ | State variables of the physical system |
| $\boldsymbol{x}$ | Spatial or spatial-temporal coordinates |
| $x$ | Spatial coordinates |
| $t$ | Temporal coordinates |
| $\theta$ | Parameters for a physical system |
| $w$ | Weights of neural networks |
| $\mathcal{F}$ | Differential operator representing the ODEs/PDEs |
| $\mathcal{I}$ | Initial conditions (operator) |
| $\mathcal{B}$ | Boundary conditions (operator) |
| $\Omega$ | Spatial or spatial-temporal domain of the system |
| $\Theta$ | Space of the parameters $\theta$ |
| $W$ | Space of weights of neural networks |
| $\mathcal{L}_{total}$ | Total loss function |
| $\mathcal{L}_{data}$ | Supervised data loss |
| $\mathcal{L}_{gov}$ | Governing equation residual loss |
| $\mathcal{L}_{BC}$ | Boundary condition loss |
| $\mathcal{L}_{IC}$ | Initial condition loss |
| $\mathcal{L}_{reg}$ | Regularization loss |
| $\|\| \cdot \|\|$ | Norm of a vector or a function |

## Appendix A

Code listings of PyTorch and DeepXDE implementation of the PINN to solve Van der Pol oscillator are compared as below.

```python
import torch
import numpy as np
from scipy.integrate import odeint

def VDP:
"""Define the ODE of VDP"""

# Define time domain and boundary conditions
t = np.linspace(0, 1.5, 1000)
y0 = [1, 0]     # y0 is the initial condition

# Call odeint in Scipy to generate numerical exact solutions
y = odeint(VDP, y0, t)

# Sample and generate the training set from the exact
 solution, including initial point (t0, y0) and endpoint
t_data=t[0:1.5:0.005]
y_data=y[0:1.5:0.005]

# Select collocation point
x_collocation = torch.linspace(0.,1.5,48).view(-1,1).
 requires_grad_(True)

class FNN:
"""Define neural networks"""

model = FNN(1, 1, 32, 3)    # Instantiate FNN

# The training process of PINN:
for i in range(24000):

# Data loss
yh = model(t_data)
Loss_data = 1 * torch.mean((yh - y_data)**2)

# Physical loss
 yhp = model(x_collocation)
 dx = torch.autograd.grad(yhp, x_collocation, torch.ones_like
  (yhp),
 create_graph=True)[0]
 dx2 = torch.autograd.grad(dx, x_collocation, torch.ones_like
  (dx),
 create_graph=True)[0]
 physics = (dx2 + 15 **2 * yhp - 5 * (1 - yhp**2) * dx)
 Loss_phys = (1e-4) * torch.mean(physics**2) * 1

 # Calculate the mean square error of the test set
 yhpp = model(t)
 mse = torch.mean((yhpp - y)**2)

 # Total loss of backpropagation
 Loss_total = Loss_data + Loss_phys
 Loss_total.backward()
```

Listing 1: Pytorch Implementation of PINN to Solve VDP

```python
import deepxde as dde
import numpy as np
from scipy.integrate import odeint

def VDP:
"""Define the ODE of VDP"""

# Call odeint in Scipy to generate numerical exact solutions
y = odeint(VDP)

# Sample and generate the training set from the exact
 solution, including initial point (t0, y0) and endpoint
t_data=t[0:1.5:0.005]
y_data=y[0:1.5:0.005]

geom = dde.geometry.TimeDomain(0.0, 1.5) # Define time
 domain

def boundary(t, on_boundary):
return on_boundary and dde.utils.isclose(t[0], 0)

 def error_derivative(inputs, outputs, X):
 return dde.grad.jacobian(outputs, inputs, i=0, j=None)

 # Define initial conditions: y(0)=1 and y'(0)=0
 ic1 = dde.icbc.IC(geom, lambda x : 1, lambda _, on_initial:
  on_initial)
 ic2 = dde.icbc.OperatorBC(geom, error_derivative, boundary)

 # Define training set and boundary conditions (including
  boundary conditions)
 t_y_data = dde.PointSetBC(t_data, y_data)

 data = dde.data.TimePDE(geom, ode, [ic1, ic2, t_y_data],
  num_domain=40, num_boundary=1)    # num_domain is the
  number of collocation points

 net = dde.maps.FNN([1] + [32] * 3 + [1])   # Instantiate FNN
 model = dde.Model(data, net)
 model.train(epochs=24000)
```

Listing 2: DeepXDE Implementation of PINN to Solve VDP